\documentclass[journal]{IEEEtran}
\usepackage{amsmath}
\usepackage{amsfonts}
\usepackage{algcompatible}
\usepackage{algorithm}
\usepackage{array}
\usepackage{textcomp}
\usepackage{stfloats}
\usepackage{url}
\usepackage{verbatim}
\usepackage{graphicx}
\usepackage{cite}
\usepackage{amssymb}
\usepackage{subcaption}
\usepackage[table]{xcolor}
\usepackage{multirow}
\usepackage{booktabs}
\usepackage{mathtools}
\usepackage{tablefootnote}
\usepackage{fontawesome}
\usepackage{hyperref}
\usepackage{orcidlink}

\newcommand{\ra}[1]{\renewcommand{\arraystretch}{#1}}

\begin{document}

\title{Robo-Platform: A Robotic System for Recording Sensors and Controlling Robots}

\author{Masoud Dayani Najafabadi \orcidlink{0000-0000-0000-0000}, Khoshanm Shojaei \orcidlink{0000-0000-0000-0000} 
\thanks{The authors are with the Department of Electrical Engineering, Islamic Azad University, Najafabad Branch, Isfahan, Iran.}}

\maketitle

\begin{abstract}
Mobile smartphones compactly provide sensors such as cameras, IMUs, GNSS measurement units, and wireless and wired communication channels required for robotics projects. They are affordable, portable, and programmable, which makes them ideal for testing, data acquisition, controlling mobile robots, and many other robotic applications. A robotic system is proposed in this paper, consisting of an Android phone, a microcontroller board attached to the phone via USB, and a remote wireless controller station. In the data acquisition mode, the Android device can record a dataset of a diverse configuration of multiple cameras, IMUs, GNSS units, and external USB ADC channels in the rawest format used for, but not limited to, pose estimation and scene reconstruction applications. In robot control mode, the Android phone, a microcontroller board, and other peripherals constitute the mobile or stationary robotic system. This system is controlled using a remote server connected over Wi-Fi or Bluetooth. Experiments show that although the SLAM and AR applications can utilize the acquired data, the proposed system can pave the way for more advanced algorithms for processing these noisy and sporadic measurements. Moreover, the characteristics of the communication media are studied, and two example robotic projects, which involve controlling a toy car and a quadcopter, are included.
\end{abstract}


\section{Introduction}

To implement novel robotics ideas, researchers, engineers, and hobbyists need to choose the appropriate hardware and software. This requirement is traditionally accomplished by embedded systems ranging from simple microcontrollers to more powerful processor-based development boards or programmable logic devices (PLDs), especially for mobile robots. These devices are usually programmed using a low-level programming language such as C/C++. The options within this domain are so varied that it is unlikely for two developers to choose the same platform, which hinders cooperation and communication of ideas.

Android smartphones provide a variety of sensors, including inertial measurement units (IMU), magnetometers, barometers, cameras, and GNSS measurement sensors, as well as communication technologies such as USB, the IEEE 802.11 protocol (Wi-Fi), Bluetooth, and Global System for Mobile Communications (GSM) in a single compact unit. The hardware is affordable and accessible to all people worldwide thanks to mass production. The Android operating system is open source, and developers usually use the Java programming language to develop applications for this platform. These features make Android devices ideal not only for robotics projects but also for sensor data acquisition.

\begin{figure}[t]
	\centering
	\includegraphics[width=.45\textwidth]{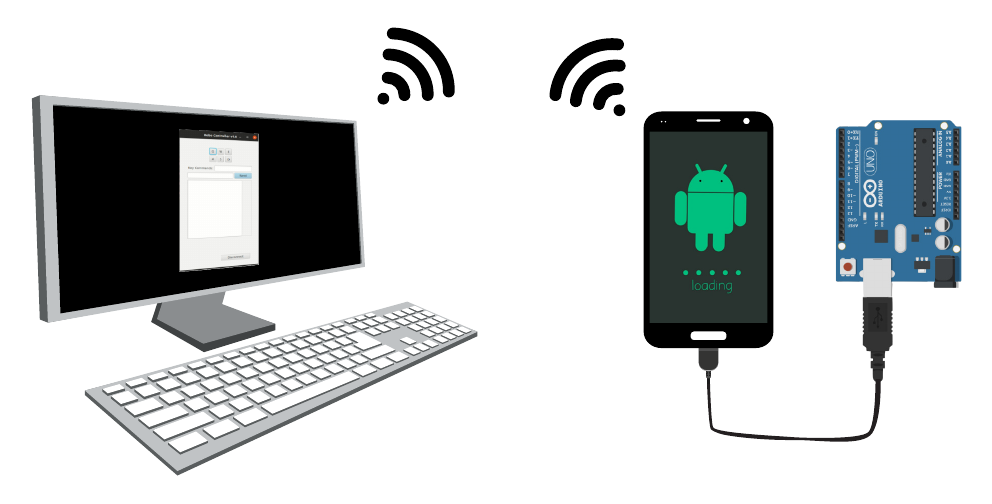}
	\caption{General configuration of the proposed system (Robo-Platform).}
	\label{fig:roboplat-system}
\end{figure}

To use Android devices for robotic applications, one should pair them with another embedded device, such as a microcontroller board, to provide general-purpose input/output (GPIO) pins needed for generating control signals. Although standard microcontroller boards such as Arduino and STM32 can power some robotic projects, they are constrained for modern autonomous applications. Another popular option is to use a more powerful processor-based embedded system, such as Raspberry Pi. However, since these modules do not provide the required sensors such as IMUs, GPS units, and cameras out of the box, they must be acquired and paired separately. Moreover, since Android phones are portable and accessible all the time, they can be paired with a commercial robot through a communication medium and control the target device. Implementing the robotic logic as a software package on general-purpose devices such as Android phones provides flexibility; for instance, an autonomous car controller can be replaced by a flight controller for an aerial vehicle.

Furthermore, data from Android sensors can be recorded for various purposes, from the medical study of patients' physical habits to the offline assessment of the dynamics of industrial and robotic systems. This feature is desired particularly in cases where a dedicated data acquisition system is costly or too complex and requires special training. Another application of sensor data acquisition is 3D scene reconstruction and pose estimation using simultaneous localization and mapping (SLAM) methods for AR/VR projects. Although mobile phones are equipped with various sensors, the collected data is often distorted and unpredictable, making it challenging for robust and accurate systems. A data acquisition system for these platforms can benefit the research community in studying such data and developing efficacious algorithms to address these issues.

The main goal of this work is to make robotics accessible to anyone worldwide by providing a low-cost robotic and data acquisition system with a modular design. In its most general form, the proposed system consists of an Android application to facilitate data acquisition and provide an interface between a remote controller server and a microcontroller board used for low-level signal generation (Fig. \ref{fig:roboplat-system}). The main contributions of this work are:

\begin{itemize}
	\item One or more built-in accelerometers, gyroscopes, magnetometers, GNSS units, cameras, and various ADC channels from an external USB microcontroller device can collect sensor data in the rawest format, suitable for scene reconstruction and pose estimation algorithms.
	\item Support for the Raw GNSS and various physical and logical camera configurations are provided.
	\item Sensor calibration data can also be recorded.
	\item A desktop or mobile application enables the remote control of the client's robotic program.
	\item Two examples of robotic clients for manual control of a toy car and a quadcopter.
	\item Designs for a compact and affordable AVR-based microcontroller board that can be paired with the Android's robotic client to provide GPIO are proposed.
	\item Benchmarking tools and post-processing scripts are provided to assess system performance and prepare the acquired data to be processed by SLAM algorithms.
	\item A sample small dataset of the recorded Android sensors is also available%
	\footnote{\href{https://drive.google.com/drive/folders/1OZqdA1xa-SyJ64qL_TibqhtwhR1fWWrx?usp=sharing}{The Robo-Platform Dataset}}.
	\item All software programs are open-sourced and provided in addition to hardware schematics and PCB designs for the benefit of the community%
	\footnote{\url{https://github.com/m-dayani/robo-platform}}.
\end{itemize}

The following section provides an overview of related works in mobile data acquisition and robot control using mobile phones. Section \ref{sec:algorithm} presents an in-depth description of the proposed system. The experiments are discussed in Section \ref{sec:experiments}. Finally, Section \ref{sec:conclusion} concludes this paper and provides future directions.

\section{Related Work} \label{sec:related:work}


\subsection{Data Acquisition}

Data acquisition with handheld mobile devices is available in many commercial and scientific applications. The purpose of data acquisition, performance, efficient resource management, modular and extensible design, code access, and application security are some of the most important aspects that should be considered when developing sensor-recording mobile applications.

Physics Toolbox Suit%
\footnote{\href{https://play.google.com/store/apps/details?id=com.chrystianvieyra.physicstoolboxsuite&hl=en_US}{The Physics Toolbox Suit}}
is a commercial application that allows users to visualize and record a database of a range of internal sensors, from the IMU to the proximity sensor. The current version of this program at the time of this writing cannot record data from the phone's cameras, microphone, and externally attached sensors. Google's GnssLogger%
\footnote{\url{https://github.com/google/gps-measurement-tools}}
only supports newer devices and allows the recording and visualization of detailed raw GNSS information.

GetSensorData \cite{get-sensor-data} is a modular and open-source Android application for visualizing a wide range of internal and external sensors, including the phone's camera and logging sensors that output textual data (most sensors except the camera) in a single file. Although the GetSensorData and Robo-Platform are mainly developed for pose and state estimation, the proposed system provides an organized folder and data structure for recording the phone's sensors and cameras in the rawest format. Also, the current version of Robo-Platform can only record data from a microcontroller via USB and provides no live data except for the camera preview.

Open Data Kit (ODK) Sensors \cite{odk-sensors} provided a modular application-level driver to simplify data acquisition for non-expert developers. Unlike Robo-Platform, which mainly focuses on the phone's internal sensors, ODK Sensors deals with external sensors used in medical applications and attached via USB. Although Robo-Platform is not optimized for performance and does not separate the logic for communication channels from sensors, it implements an efficient internal message-passing system that allows developers to register data listeners and acquire sensor measurements.

Similar to the proposed system, the Mobile AR Sensor (MARS) Logger \cite{mars-logger} can record accurate calibration information and raw sensor data required for state estimation in SLAM and AR applications. However, while they record frames as a video and try to calibrate the time offset between the frames and IMU readings offline, the proposed method can record uncalibrated data from multiple accelerometers, gyroscopes, magnetometers, GNSS, timestamped images from one or more physical and logical cameras, and external sensors via the USB. Hence, the format of the recorded dataset is compatible with popular SLAM datasets such as the EuRoC MAV \cite{euroc-dataset} and TUM RGB-D \cite{tum_rgbd_sturm12iros} datasets. Although MARS Logger supports Android and iOS platforms, the current version of Robo-Platform is only developed for Android devices.

ADVIO \cite{advio_dataset} presented a public benchmark dataset recorded using several devices for visual-inertial odometry and SLAM and a method for generating ground truth from the data. More specifically, the accelerometer, gyroscope, magnetometer, barometer, GNSS data, and video footage were recorded in addition to the computed ground truth using an Apple iPhone 6S phone, which was reported to be a mid-range consumer device. Sensor data were temporally synchronized using the network time protocol (NTP) requests. The recorded video has a much higher frame rate and resolution, but its portrait orientation limits the field of view, which is problematic for most VI-SLAM methods. Moreover, this method cannot record data from multiple sensors of the same type and has no support for external sensors.

Finally, VINS-Mobile%
\footnote{\url{https://github.com/HKUST-Aerial-Robotics/VINS-Mobile}}
for iOS and VINS-Mobile-Android%
\footnote{\url{https://github.com/jannismoeller/VINS-Mobile-Android}}
for Android are two applications that implement VINS-Mono \cite{vins_mono_2018} for mobile devices. PTAM \cite{ptam_2007} was developed to enable AR on handheld devices. Although these pipelines does not record sensor data for later use, they capture visual and inertial data for SLAM and AR applications.


\subsection{Robot Control}

\begin{figure*}[t]
	\centering
	\includegraphics[width=.60\textwidth]{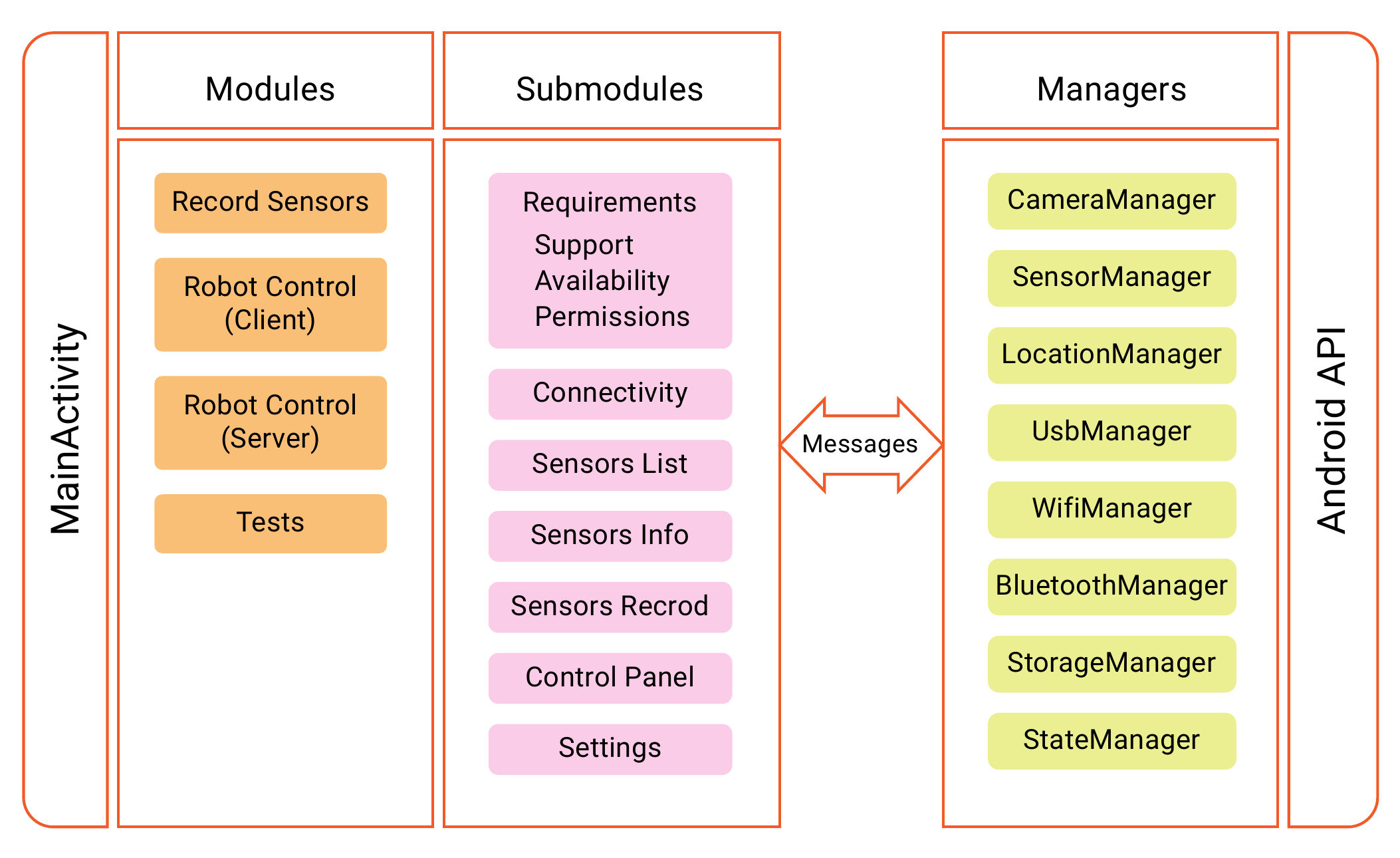}
	\caption{General architecture of the Robo-Platform Android application.}
	\label{fig:roboplat-android-arch}
\end{figure*}

Most related robotics applications utilize various communication protocols (Wi-Fi, Bluetooth, USB) to interface an Android phone to a proprietary robot. This interface usually requires transmitting commands specific to that robotic platform, which restricts the system's adoption to other applications. In this respect, an architecture that transmits general messages to an intermediate medium with adequate signaling inputs and outputs (e.g., an Arduino board) and leaves the interpretation of the commands to it is preferred.

An application for monitoring and controlling a remote commercial industrial robot was presented in \cite{monitor_control_industrial_robot}, similar to Robo-Platform in server mode. The Android application connects to a proprietary control program via Wi-Fi and can send commands and receive and monitor the robot's state. Although Robo-Platform is designed mainly for mobile robots and does not support monitoring, it supports both Wi-Fi and Bluetooth technologies, and its text-based interface allows it to interact with any custom robot. 
Moreover, the proposed method can also capture a variety of internal and external sensors essential for most robotic applications.

Azeta et al. \cite{android_robot_surveillance} proposed a robotic system consisting of a smartphone, an Arduino-powered ground vehicle, and wireless communication through a Wi-Fi module for monitoring, surveillance, and obstacle avoidance. The user interacts with the Android phone via the wireless channel, which communicates with the Arduino to control the robot and send live camera feedback to the user. In contrast, the Robo-Platform client controller can communicate with a desktop or mobile application via Wi-Fi or Bluetooth and interact with Arduino and an affordable custom-made microcontroller board using USB. The current version of Robo-Platform does not support sending a camera feed for monitoring.


An Android application that can safely operate a remote and custom-made robotic arm based on Android's motion and proximity sensors was presented in \cite{salah2024smartphone}. The Android phone sends sensor-based commands to an Arduino microcontroller board via Bluetooth, and the Arduino generates the necessary servo motors' signals to manipulate the robotic arm. This project presented an example of capturing Android's sensors and utilizing the obtained data for robot control. Although the Robo-Platform in the client mode can use IMU data to stabilize the flight control operation, the current version of the proposed system does not use sensor data in the server mode for robot control.


\section{Algorithm} \label{sec:algorithm}

Fig. \ref{fig:roboplat-system} shows a general configuration of the Robo-Platform system. The proposed system consists of three components. The desktop server wirelessly sends control commands and data to the remote Android sub-module. The Android application processes the commands and controls the microcontroller client using the USB protocol. The microcontroller board provides additional sensors and GPIO pins and generates low-level control signals.

Fig. \ref{fig:roboplat-android-arch} illustrates the general layout of the proposed Android module. The entry point of this application is the \emph{Main Activity}, which represents and provides access to the main tasks. Each task is an independent application (Android \emph{Activity}), which usually manages and interacts with several sub-modules (Android \emph{Fragment}). These sub-modules interact with \emph{Managers} through an internal efficient message-passing system to accomplish their purpose. Managers are responsible for most core tasks and usually wrap one or more Android API functionality.

The current version of the Android application operates in two main modes: Sensor Recording and Robot Control. In the former, the user selects from a list of available sensors and proceeds to the recording step when satisfied with the selection. In the robot controller mode, the user first establishes the necessary connections and transmits messages and commands between a remote desktop server and a microcontroller platform.

\begin{figure}[ht]
	\centering
	\includegraphics[width=.40\textwidth]{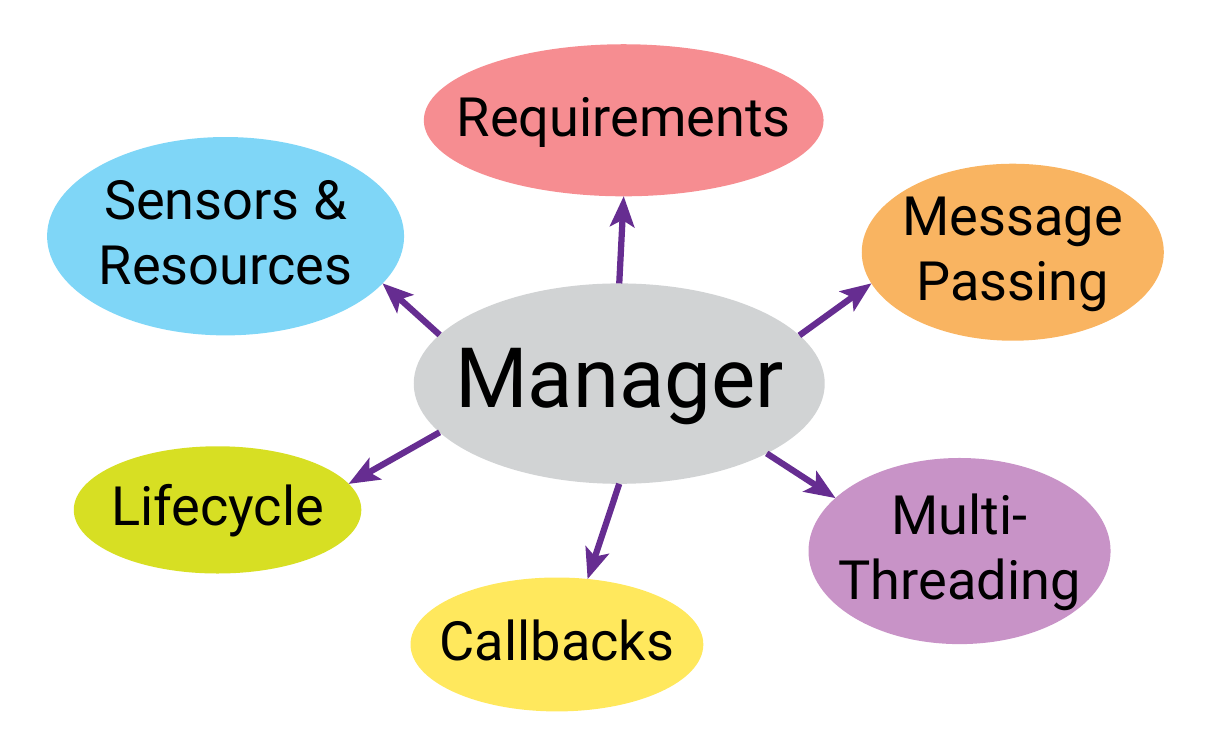}
	\caption{Essential responsibilities of Managers in the proposed Android application.}
	\label{fig:roboplat-manager-duties}
\end{figure}

The concept of sensors and managers is introduced to effectively manage the resources on an Android device. A sensor is a physical sensor in an Android-powered device like the accelerometer, external devices attached via USB, or other resources such as the storage media. Fig. \ref{fig:roboplat-manager-duties} depicts the main functions of Managers. They contain the logic to control specific sensors and handle background operations (multi-threading). To share data between different sensors, a message-passing interface enables each manager to publish sensor or configuration messages and subscribe to a publisher.

Managers also manage sensor-specific requirements. Device support for a sensor and the platform permissions are two of the most critical requirements. In addition, the user must enable some features before access. In the case of communication sensors, it is essential to establish a reliable communication medium and ensure that the remote device responds correctly to the configuration messages.

\subsection{Recording Android Sensors} \label{subsec:sensor-recording}

\begin{figure}[ht]
	\centering
	\includegraphics[width=.45\textwidth]{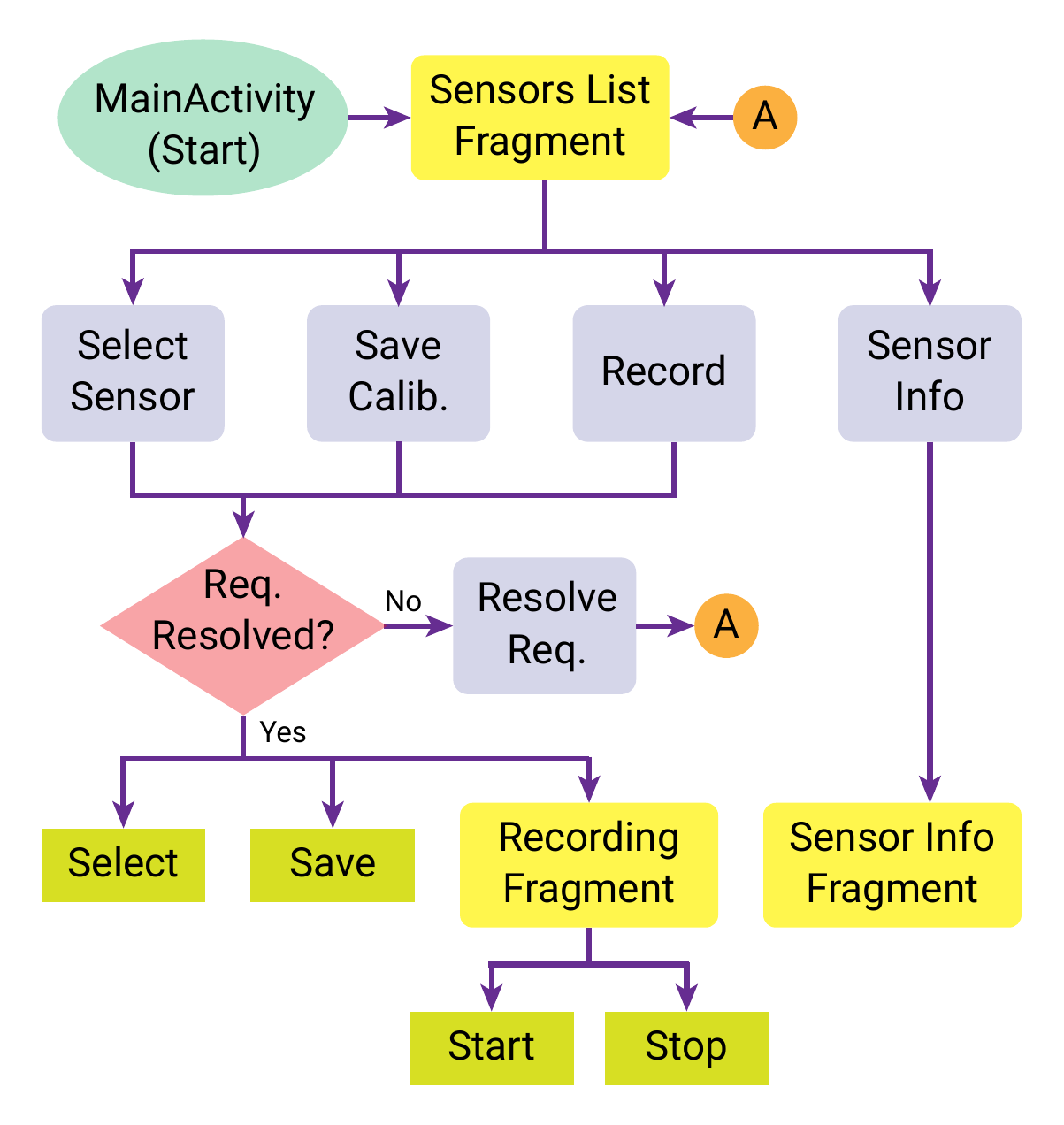}
	\caption{The flowchart of the proposed Android module in sensor recording mode.}
	\label{fig:roboplat-sensor-recording}
\end{figure}

Fig. \ref{fig:roboplat-sensor-recording} illustrates the logic of the sensor recording task. A list of supported sensors is presented to the user. They can select a sensor only after resolving all its requirements. Otherwise, they should follow the steps to resolve the permissions, enable an Android functionality, or establish a USB connection to the external sensor. The user can also inquire additional information about a specific sensor and store it if the storage permission is granted. After all necessary sensors are selected, the user can proceed to the recording screen, where they can start or stop the recording and view the activity logs and camera preview, provided that a camera sensor is selected.

\begin{table}[ht]
	\caption{A summary of the various managers, sensors, and requirements for the sensor recording task.}
	\ra{1.3}
	\begin{center}
		\begin{tabular}{@{}lll@{}}
			\hline
			Managers & Sensors/Resources & Requirements \\
			\hline 
			Sensor & IMU, Magnetic Field, etc. & Support \\
			Location & GPS, Raw GNSS & Support, Permission, Enable \\
			Camera & Timestamped Images & Permission \\
			USB & External ADC & Connectivity, Permission, Test \\
			Storage & Storage Media & Permission, Storage Path \\
			\hline 
		\end{tabular}
		\label{tab:sensor-recording-managers}
	\end{center}
\end{table}

Table \ref{tab:sensor-recording-managers} summarizes supported managers, sensors, and \emph{requirements} at this level. All operations except retrieving sensor information might depend on a set of requirements. Furthermore, saving calibration data and starting a recording session requires the availability of the storage medium. The user must resolve all the requirements before requesting a feature.

\subsubsection{IMU and Magnetic Field Sensors}

A device might have several accelerometers, gyroscopes, and magnetometers. Each type of sensor is stored in a separate file. IMU (accelerometers and gyroscopes) and magnetic field sensors are considered separate sensor groups, and their data are stored in different folders (Fig. \ref{fig:roboplat-folder-structure}). The application can record pre-calibrated and raw, uncalibrated measurements. These sensors usually do not have any requirements. Other types of sensors, such as barometers, are not considered in the current version of the algorithm.

\subsubsection{GNSS}

The Android platform initially provided GPS measurements as the source of location data. These measurements are stored in a specific file. However, raw GNSS has also been available in recent versions. For the raw GNSS readings, both Navigation and Measurement Messages are stored in separate files to allow the recovery of the device's geographical location.

\subsubsection{Camera}

Most portable devices provide multiple cameras. Additionally, each physical camera can provide different output streams, e.g., one for preview and another for image processing and storage. According to the Android Camera2 API \cite{camera2-api-doc}, a logical camera in a device that supports multiple camera outputs represents several physical cameras. In this case, the user can select any combination of physical cameras belonging to a specific logical camera group but not from the other groups.

Instead of saving a video stream for each physical camera, timestamped images are stored in a designated folder. A separate text file containing all image names and their corresponding timestamps is also stored. Since most devices are equipped with auto-calibration algorithms, a locking algorithm is triggered at the start of the recording session to lock the focal length and other parameters (white balance and exposure). Moreover, if the device supports lens distortion correction, this capability is also disabled for capturing high-rate raw images. Available parameters can be stored in the calibration file and retrieved for post-processing.

\subsubsection{External USB Device}

The platform also supports recording external ADC channels from a microcontroller via the USB. The microcontroller fills its output buffer when an ADC channel reports a new value. Then, the Android platform initiates an ADC report command, receives the ADC measurements as a raw array of bytes, and publishes the result, timestamped with the received time. Therefore, the data acquisition rate depends on the communication latency and the microcontroller's configured sample rate.

The microcontroller device can also store the calibration data, such as the number of ADC channels, resolution in bits, and the sample rate. The Android platform can retrieve and store the calibration information after establishing a communication channel.

\subsubsection{Folder Structure}

\begin{figure}[ht]
	\centering
	\includegraphics[width=.22\textwidth]{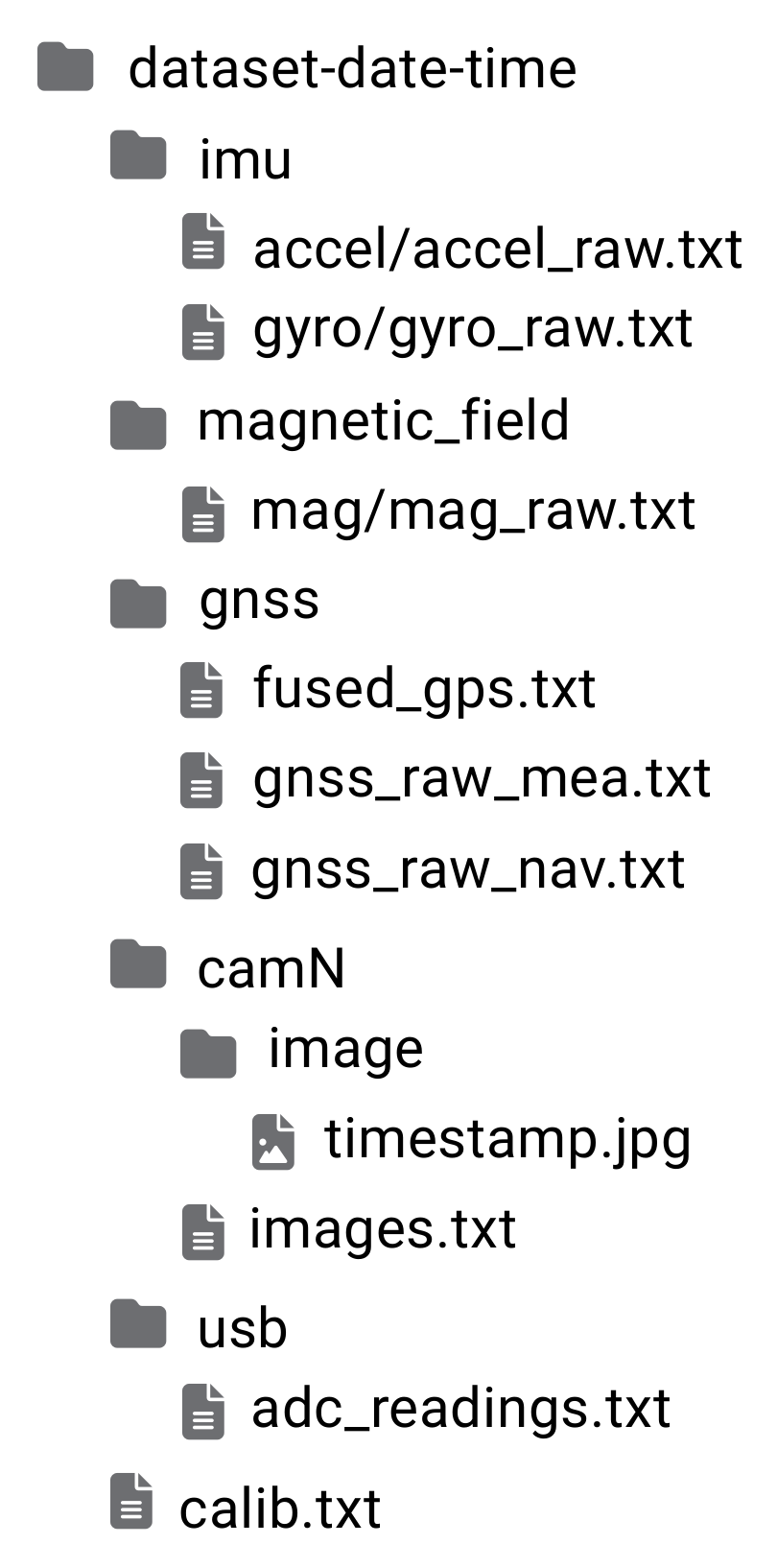}
	\caption{File and folder structure of the recorded dataset.}
	\label{fig:roboplat-folder-structure}
\end{figure}

Fig. \ref{fig:roboplat-folder-structure} shows the proposed folder structure for the recorded dataset. Each recording session creates a root dataset folder. Measurements from related sensors are stored in separate files in the same folder. The raw GNSS sensor records its observations in multiple files. Each physical camera stores a text file for timestamped image paths and a folder of all images. Each row of the text files that store readings from multiple sensors of the same type also includes a sensor identifier.

\subsubsection{Data formats}

\begin{table*}[t]
	\caption{The stored data format for each type of sensor.}
	\ra{1.3}
	\begin{center}
		\begin{tabular}{@{}ll@{}}
			\hline
			Sensor & Format \\
			\hline 
			Gyroscope & timestamp\_ns, rx\_rad\_s, ry\_rad\_s, rz\_rad\_s, [b\_rx\_rad\_s, b\_ry\_rad\_s, b\_rz\_rad\_s,] sensor\_id \\
			Accelerometer & timestamp\_ns, ax\_m\_s2, ay\_m\_s2, az\_m\_s2, [b\_ax\_m\_s2, b\_ay\_m\_s2, b\_az\_m\_s2,] sensor\_id \\
			Magnetic Field & timestamp\_ns, mx\_uT, my\_uT, mz\_uT, [b\_mx\_uT, b\_my\_uT, b\_mz\_uT,] sensor\_id \\
			GPS & timestamp\_ns, latitude\_deg, longitude\_deg, altitude\_m, velocity\_mps, bearing \\
			GNSS Navigation & timestamp\_ns, sv\_id, nav\_type, msg\_id, sub\_msg\_id, data\_bytes\_hex \\
			GNSS Measurement & timestamp\_ns, time\_offset\_ns, rx\_sv\_time\_ns, acc\_delta\_range\_m, ps\_range\_rate\_mps, cn0\_DbHz, snr\_db, \\
			 & cr\_freq\_hz, cr\_cycles, cr\_phase, sv\_id, const\_type, [bias\_inter\_signal\_ns, type\_code] \\
			Camera & timestamp\_ns, image\_path.jpg \\
			External ADC & timestamp\_ns, adc\_reading, [adc\_channel\_id] \\
			\hline 
		\end{tabular}
		\label{tab:sensor-data-format}
	\end{center}
\end{table*}

Table \ref{tab:sensor-data-format} summarizes the data format for each type of sensor. 
All measurements are stamped with the time the sensor value becomes available with nanosecond accuracy. All text files also contain a header describing the content of the file. Some data must be interpreted by the user. For example, what ADC values represent depends on the physical quantities the external sensors measure.

In Table \ref{tab:sensor-data-format}, the square bracket means the fields are optional. In the case of raw acceleration ($\frac{m}{s^2}$), angular velocity ($\frac{rad}{s}$), and magnetic field ($\mu T$), the first three columns are raw measurements before the optional sensor biases. For a comprehensive explanation of raw GNSS fields, please refer to \cite{smartphone-decimeter-2023}.

\subsection{Robot Controller} \label{subsec:robot-controller}

The robot controller can operate in either server or client mode. Each mode establishes a wireless communication channel using Wi-Fi or Bluetooth technologies.

\begin{figure}[ht]
	\centering
	\includegraphics[width=.45\textwidth]{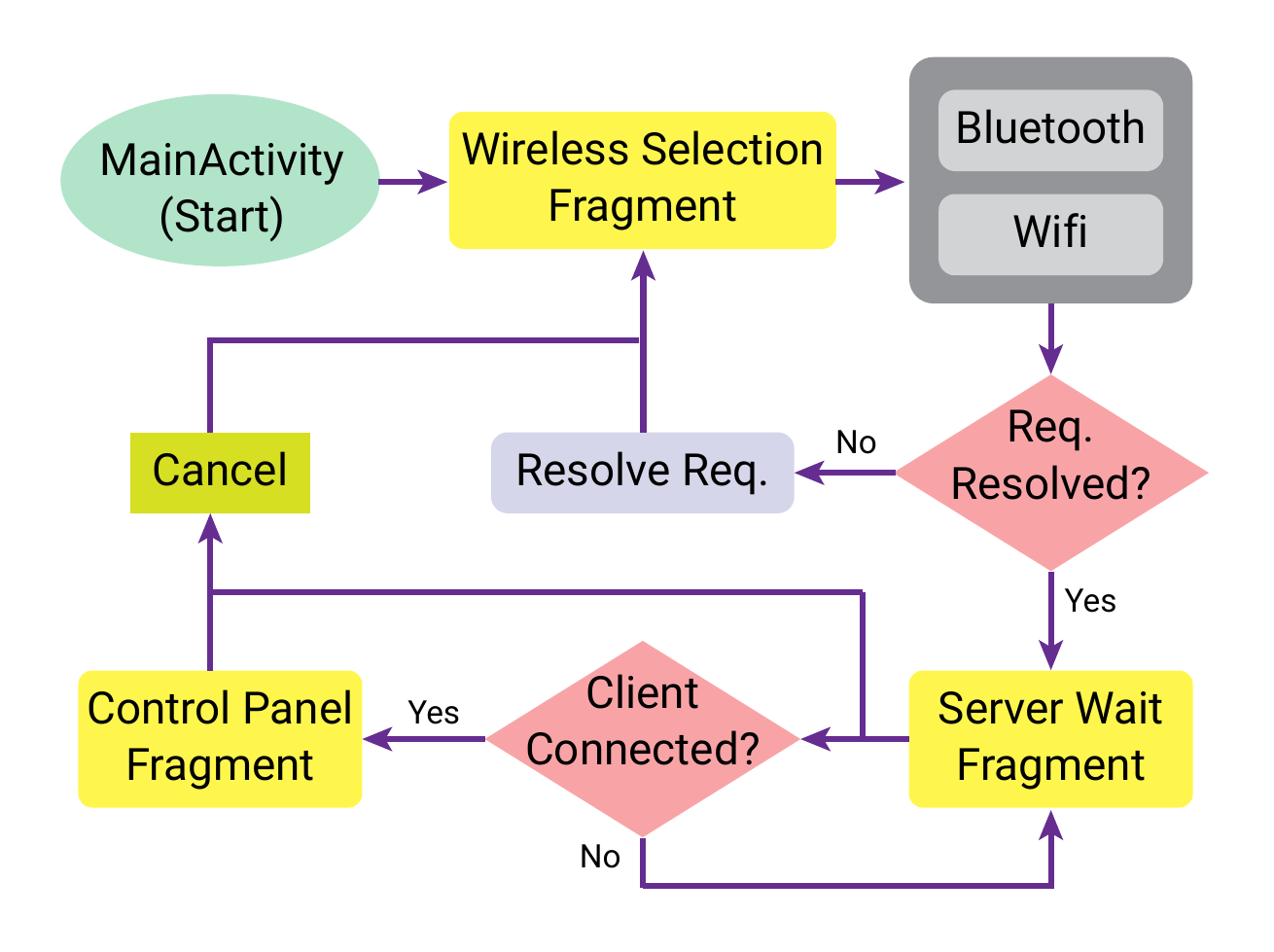}
	\caption{Robot controller server logic.}
	\label{fig:roboplat-rc-server}
\end{figure}

Fig. \ref{fig:roboplat-rc-server} shows the robot controller server flowchart. The user first selects the communication mode. The application then checks if the selected mode is available and listens over a defined port for incoming connection requests. When the client requests a communication channel, the server accepts the request, and a control panel appears on the screen.

\begin{figure}[ht]
	\centering
	\includegraphics[width=.45\textwidth]{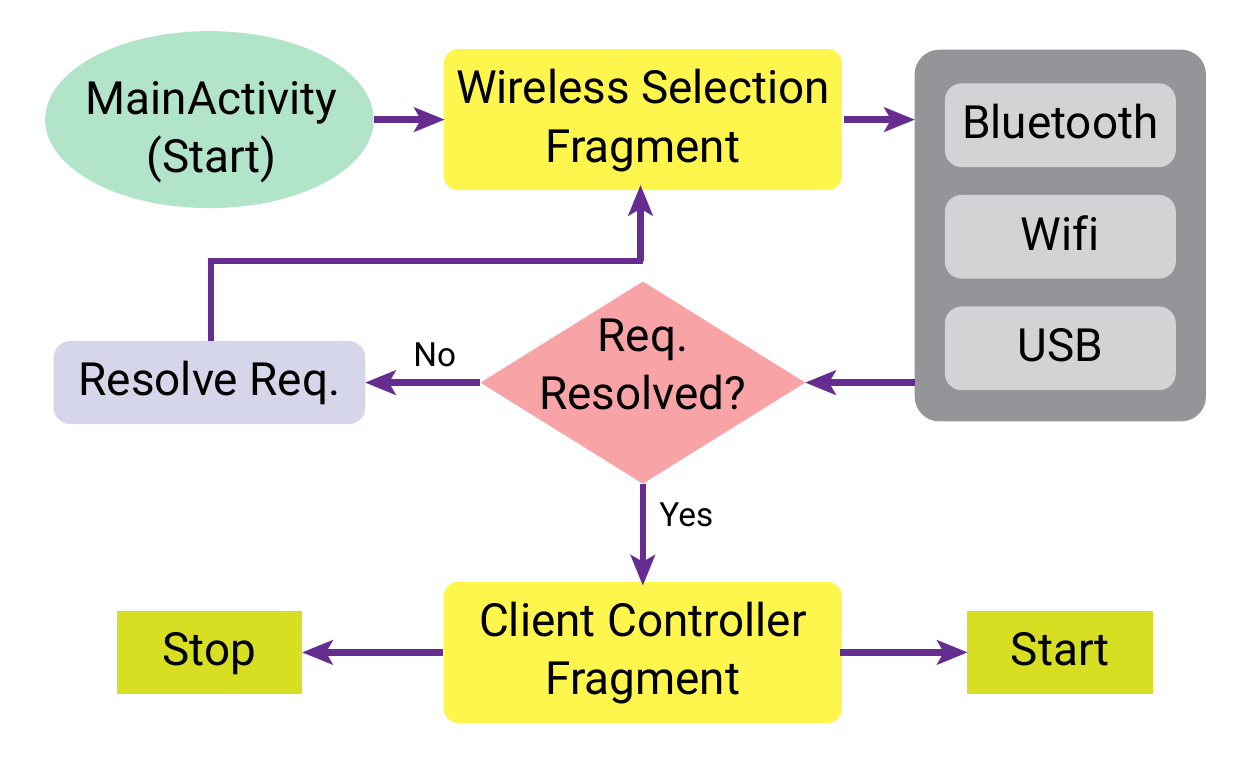}
	\caption{Robot controller client logic.}
	\label{fig:roboplat-rc-client}
\end{figure}

On the client side (Fig. \ref{fig:roboplat-rc-client}), the user first enters the server's address. The server address in Wi-Fi and Bluetooth modes is indicated by the IP address/port number and the paired server's name/port, respectively. Additionally, the client should establish a USB connection to the microcontroller board. In this case, the Android device acts as the server of the USB channel but the client of the wireless connection. After establishing all connections, the platform can receive commands from the remote desktop and publish them to the microcontroller.

A test message is sent to the client to ensure the communication nodes belong to the platform. Subsequently, the response message is received and checked for validation. Moreover, we can implement an encryption mechanism to safeguard communications from unauthorized exposure.

\subsubsection{Desktop Program}

In addition to the Android wireless server, a desktop controller server is presented. The logic is the same as the Android server explained before.

\subsubsection{AVR Device}

In addition to the Arduino microcontroller board, the proposed system can communicate with a custom AVR board through the USB channel. There are several hardware and software USB implementations for AVR boards. The proposed design is powered by the OBDEV's V-USB software library \cite{obdev-v-usb-link}. The AVR module receives USB commands, decodes them, and executes the requested task. It can also load buffers with data (ADC sensor and configuration) and send them to the mobile application on demand. This software-based implementation can be very cost-effective since it obviates the need for an external USB controller chip and extra components.

Another popular option is to use third-party embedded boards such as an Arduino board. In this case, the USB communication incorporates Arduino's standard serial transfer.

\section{Experiments} \label{sec:experiments}

The Android module of the proposed system is developed using the Java programming language for the newest API (API 35 at the time of this writing) with a minimum API level of 21, which provides support for approximately 99.7\% of all commercial devices. The desktop server is also developed in Java and uses JavaFX for the user interface, as well as the Bluecov library for Bluetooth connectivity.


The proposed method supports a range of Arduino boards and a custom AVR board. A USB client logic has been implemented using the C programming language and the V-USB library for two Atmel AVR microcontrollers (ATmega8 and ATmega32). The code, board schematics, and PCB designs are provided on the project page.


\subsection{Record Datasets}

\begin{table*}[th]
	\caption{A description of sequences and datasets used to study the performance of various state estimation algorithms (\emph{Mag.} stands for \emph{Magnetometer}).}
	\ra{1.3}
	\begin{center}
		\begin{tabular}{@{}lccccll@{}}
			\hline
			Dataset & Camera & IMU & Mag. & GPS & Other & Comments \\
			\hline 
			EuRoC \cite{euroc-dataset} & {\tiny\faCircle} & {\tiny\faCircle} & & & Leica laser tracker, Vicon motion capture & $752\times480$ stereo images \\
			TUM-VI \cite{tum_vi_dataset} & {\tiny\faCircle} & {\tiny\faCircle} & & & OptiTrack motion capture & Square 512 \& 1024 pixel stereo images \\
			Public Event \cite{ethz_pub_ds_Mueggler_2017} & {\tiny\faCircle} & {\tiny\faCircle} & & & Events & $240\times180$ monocular DAVIS images \\
			Complex Urban \cite{complex_urban_ds} & {\tiny\faCircle} & {\tiny\faCircle} & {\tiny\faCircle} & {\tiny\faCircle} & 2D \& 3D LiDAR, FOG, Altimeter, Encoder & $1280\times560$ stereo images \\
			ADVIO \cite{advio_dataset} & {\tiny\faCircle} & {\tiny\faCircle} & {\tiny\faCircle} & {\tiny\faCircle} & Barometer, ARKit, Google ARCore \& Tango & $720\times1280$ monocular iPhone video \\
			Robo-Plat. (proposed) & {\tiny\faCircle} & {\tiny\faCircle} & {\tiny\faCircle} & {\tiny\faCircle} & Raw GNSS, Ext. USB & $640\times480$ monocular images \\
			\hline 
		\end{tabular}
		\label{tab:dsFeaturesSummary}
	\end{center}
\end{table*}

This section first reviews the data acquisition statistics. Next, several datasets recorded using the proposed system are processed with sample state estimation algorithms and compared against similar datasets regarding the estimation duration.

\subsubsection{Data Acquisition Statistics}

\begin{table}[htb]
	\caption{Data acquisition statistics for the recorded samples using the proposed platform.}
	\ra{1.3}
	\begin{center}
		\begin{tabular}{@{}lllll@{}}
			\hline
			Sensor & Mean Period & Period STD & Number of & Duration \\
			& (s) & (s) & Samples (k) & (s) \\
			\hline 
			Gyro & 0.01 & 0.002 & 34 & 349 \\
			Gyro Raw & 0.01 & 0.002 & 551 & 5580 \\
			Accelerometer & 0.01 & 0.002 & 580 & 5886 \\
			Magnetometer & 0.01 & 0.003 & 4 & 43 \\
			Mag. Raw & 0.01 & 0.002 & 551 & 5580 \\
			Ext. ADC & 0.01 & 0.003 & 1 & 14 \\ 
			Camera & 0.32 & 0.038 & 8 & 2510 \\
			GPS & 4.06 & 0.378 & 1 & 4276 \\
			\hline 
		\end{tabular}
		\label{tab:daq-stats}
	\end{center}
\end{table}

Sample datasets are recorded using various devices and sensor configurations. Then, the average and standard deviation of each sensor's period is calculated using the timestamps of all similar measurement files and reported in Table \ref{tab:daq-stats}. According to the results, most motion sensors can output data at about 100 Hz with acceptable accuracy (relatively small jitter). The external ADC can also operate at 100 Hz, which is beneficial in capturing rapid fluctuations in the measured quantity. The camera in this experiment can capture about three frames per second, much lower than a video captured at 30 fps. However, recording in this format is more reliable and necessary for some applications such as AR. Finally, GPS data exhibit the lowest frequency (approximately 0.25 Hz), which is expected with consumer GNSS sensors.

\subsubsection{Sample Datasets for Pose Estimation}

This section aims to answer the question, can modern state estimation methods process the noisy and sporadic sensor data from a mobile phone to reconstruct the scene and estimate the device pose? Several datasets are captured using the proposed method and fed into four visual and inertial state estimation algorithms. This process is repeated for other related datasets, and the tracking duration is calculated and reported as the percentage of the successful tracking time to the duration of a whole sequence.

ORB-SLAM3 \cite{orbslam3_2021} is a robust SLAM algorithm that supports several sensor configurations and camera models. The monocular visual and visual-inertial settings of this algorithm are considered in this experiment. Another popular state estimator is VINS-Fusion \cite{vins_fusion_2019}, used in the monocular visual-inertial configuration. Finally, Matlab's asynchronous Extended Kalman Filter (EKF) state estimator \cite{matlab_insFiltAsync} is used to fuse measurements from the accelerometer, gyroscope, magnetometer, and GPS sensors and estimate the pose of the device when no visual data is available.

\begin{figure}[htbp]
	\centering
	\begin{subfigure}{.25\textwidth}
		\centering
		\includegraphics[width=.85\textwidth]{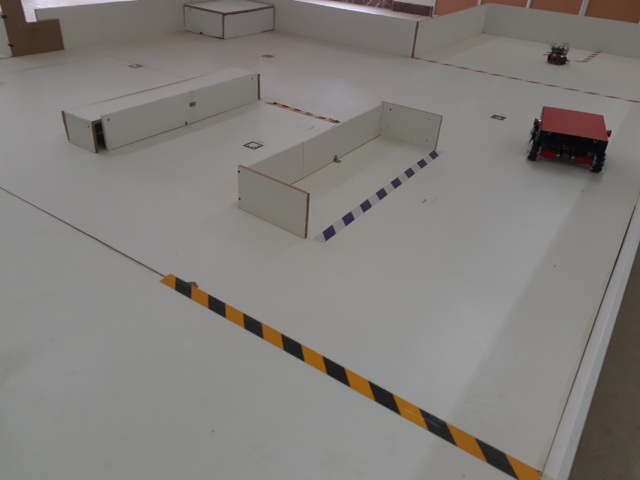}
		\caption{}
		\label{fig:sampleDsImages:iaunCvrLab}
	\end{subfigure}%
	\begin{subfigure}{.25\textwidth}
		\centering
		\includegraphics[width=.85\textwidth]{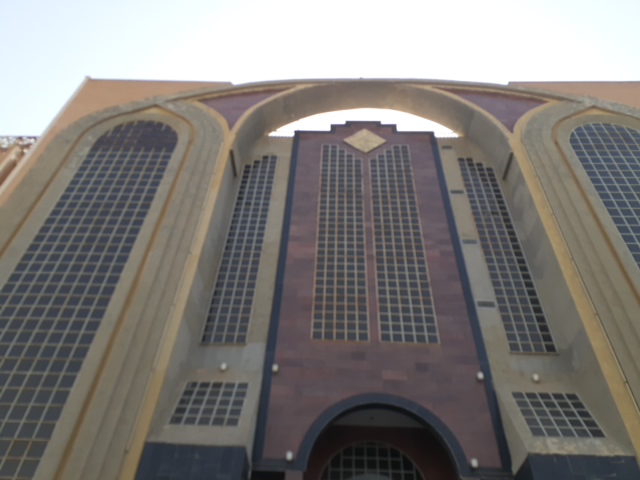}
		\caption{}
		\label{fig:sampleDsImages:iaunLib}
	\end{subfigure}
	\begin{subfigure}{0.25\textwidth}
		\centering
		\includegraphics[width=.85\textwidth]{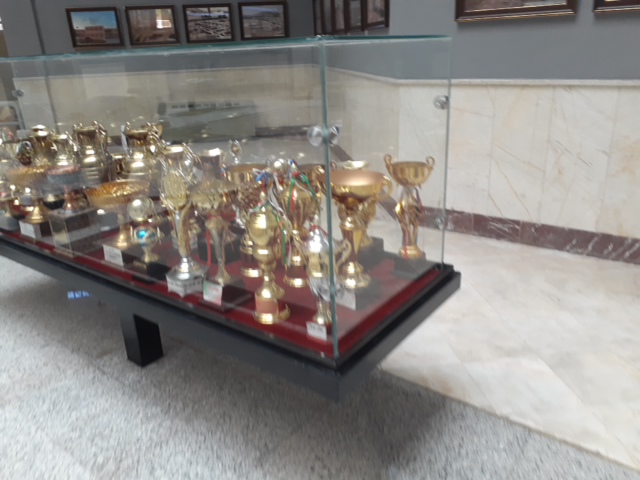}
		\caption{}
		\label{fig:sampleDsImages:iaunCorridor2}
	\end{subfigure}%
	\begin{subfigure}{0.25\textwidth}
		\centering
		\includegraphics[width=.85\textwidth,trim={0cm 0.7cm 0cm 0.7cm},clip]{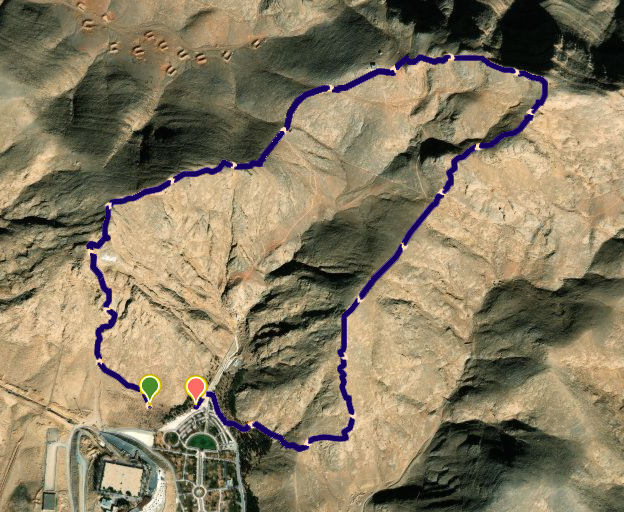}
		\caption{}
		\label{fig:sampleDsImages:koohestanPark}
	\end{subfigure}
	\caption{%
		Sample images of the recorded dataset using the proposed Android application. The sequences are: 		(\subref{fig:sampleDsImages:iaunCvrLab}) iaun-cvr-lab, (\subref{fig:sampleDsImages:iaunLib}) iaun-library, (\subref{fig:sampleDsImages:iaunCorridor2}) iaun-corridor2, and (\subref{fig:sampleDsImages:koohestanPark}) koohestan-park (track).%
	}
	\label{fig:sampleDsImages}
\end{figure}

EuRoC \cite{euroc-dataset} and TUM's visual inertial dataset \cite{tum_vi_dataset} are popular SLAM datasets and benchmarks. The reference sequences \emph{Machine Hall 3} from the former and the \emph{Corridor-4-512} from the latter are considered in this experiment. The Public Event dataset \cite{ethz_pub_ds_Mueggler_2017} provides challenging sequences recorded using a DAVIS device. Only intensity images and IMU measurements of the sequence \emph{boxes\_6dof} are used in this study. The sequence \emph{urban25-highway} from the Complex Urban Dataset \cite{complex_urban_ds} provides stereo images, inertial measurements, GPS data, and magnetic field readings. These datasets are recorded using a dedicated sensor system and provide a reference. ADVIO \cite{advio_dataset} is a relevant public dataset that records data from a configuration of multiple devices and sensors, including an iPhone mobile device. Finally, a dataset is recorded using the proposed Android application. Fig. \ref{fig:sampleDsImages} presents sample images of several sequences in this dataset. Table \ref{tab:dsFeaturesSummary} describes the datasets used in this experiment.

These state estimation algorithms require calibration parameters and a correctly formatted dataset. Calibration is the process of estimating sensor model parameters (camera intrinsics and distortion and IMU noise and bias) and the spatiotemporal relation between sensors (e.g., camera-IMU transformation and time offset). SLAM and AR datasets usually provide these parameters by calibrating their sensors or sharing a calibration dataset that can be used to calculate these parameters manually. The sensor model of a calibrated dataset must match that of the SLAM algorithm. For instance, while ORB-SLAM3 supports the distorted pinhole and fisheye \cite{kannala_brandt_cam_model} camera models, a dataset that only provides a unified projection model \cite{unified_omni_model} should be recalibrated for the correct parameters. A separate calibration dataset is recorded to estimate the calibration parameters for the proposed dataset using the Kalibr Toolbox%
\footnote{\url{https://github.com/ethz-asl/kalibr}}.

The dataset's output formats vary for different providers. Although the ROSBAG%
\footnote{\url{https://wiki.ros.org/rosbag}} 
format tries to mitigate this problem by presenting data from various sensors as standard messages passed to different components, the installation of ROS is necessary for using it. Additionally, ORB-SLAM3 supports several other directory and file structures, including that of the EuRoC dataset. For Matlab's EKF state estimator, measurements from the inertial, magnetic field, and GPS sensors must be aligned using the timestamps. Datasets recorded in unsupported formats must be converted, e.g., aligning separate accelerometer and gyroscope readings in the ADVIO and proposed datasets. Additionally, timestamped images should be extracted from the ADVIO videos. The \emph{bag\_creater} tool of the Kalibr Toolbox can create ROSBAG files from each sequence. The GitHub repository of the proposed system also provides several scripts for this post-processing step.

After these steps, the visual (V) and visual-inertial (VI) configurations of ORB-SLAM3, visual-inertial VINS-Fusion, and the inertial (I) Matlab's state estimator are executed on each sequence. Tracking time, $\tau$, is the difference between the first and last timestamps of the estimated trajectory. This time is divided by the total duration of that sequence, $T$, to calculate the percentage time of tracking. Each method-sequence combination is repeated five times, and the median percentage is reported in Table \ref{tab:ds-state-est}. A dash symbol (\emph{-}) in this table indicates a tracking failure or unsupported sensor configuration (sequence \emph{koohestan-park})%
\footnote{Video: \url{https://youtu.be/BTQ4yLB1bak}}.

\begin{table}[ht]
	\caption{The tracking time percentage of recorded datasets using the proposed platform compared against similar datasets for SLAM/AR using several state estimation algorithms.}
	\ra{1.3}
	\begin{center}
		\begin{tabular}{@{}llcccc@{}}
			\hline
			Dataset & Sequence & V \cite{orbslam3_2021} & VI \cite{orbslam3_2021} & VI \cite{vins_fusion_2019} & I \cite{matlab_insFiltAsync} \\
			\hline 
			\hline
			EuRoC & MH\_03 & 97 & 82 & 97 & 100 \\
			\hline 
			TUM-VI & corridor4 & 99 & 94 & 96 & 100 \\
			\hline 
			Public Event & boxes\_6dof & 99 & - & 98 & 100 \\ 
			\hline 
			Complex Urban & urban25 & 28 & - & 96 & 100 \\
			\hline 
			 & advio-01 & 11 & - & 97 & 100\\
			ADVIO & advio-19 & 94 & - & 99 & 100 \\
			 & advio-20 & 26 & - & 99 & 100 \\
			\hline 
			 & iaun-cvr-lab & 28 & 68 & 97 & 100 \\
			Robo- & iaun-library & 10 & - & 90 & 100 \\ 
			Platform & iaun-corridor1 & 12 & 56 & - & 100 \\
			(proposed) & iaun-corridor2 & 6 & - & 17 & 100\\
			 & koohestan-park & - & - & - & 100 \\
			\hline 
		\end{tabular}
		\label{tab:ds-state-est}
	\end{center}
\end{table}

Although ORB-SLAM3 and VINS-Fusion can track EuRoC and TUM-VI sequences with acceptable accuracy, their performance degrades in challenging conditions. VINS-Fusion and inertial EKF can follow almost all sequences with a high percentage. However, the estimation error always overgrows in the latter case, and the same is true for the VINS-Fusion in challenging situations. This issue is particularly severe for purely IMU-based sequences because, without the magnetic field and GPS measurements, inertial EKF fails to correct IMU errors. The accuracy of the methods is not reported here because it assesses the limitations of algorithms rather than datasets.

The visual-inertial ORB-SLAM3 fails to track most sequences, while its visual configuration shows poor performance for the challenging sequences. Interestingly, VI ORB-SLAM can better follow the sequences recorded by the proposed platform. The sequence \emph{advio-19} shows a sudden jump in the V ORB-SLAM results because it fuses a few loop-closing poses at the end of the sequence with the initial estimates.

Tracking results strongly depends on the limitations of each method. The optical-flow feature tracking and association pipeline of VINS-Fusion is sensitive to fast-paced motion, which happens frequently in the low frame rate of the proposed sequences. On the other hand, although the descriptor-based feature-tracking ORB-SLAM3 can cope with high baseline images, it is sensitive to lighting and illumination changes in the environment, which is the case in the sequence \emph{iaun-corridor4}. Current SLAM algorithms are optimized for several datasets recorded with high-quality sensors and optimal calibration; however, their performance degrades rapidly with noisy data and calibration parameters. The reverse is also true: knowing the limitations of a SLAM algorithm, it is possible to record an ideal dataset that favors a particular method.

Based on this discussion, we need robust general state estimation algorithms capable of processing low-quality data without requiring the user to go through the complex calibration procedure. The optimal hybrid method is a delicate balance of simple, low-precision, fast algorithms such as an EKF for high-frequency and noisy data and accurate but slower methods for processing sporadic and feature-rich data such as images.


\subsection{Control Mobile Robots}

An essential aspect of a robotic system is the reliable communication link between its modules. Communication bandwidth, latency, throughput, and range are important factors when considering a communication link's performance. Bandwidth and range depend on the environment and the real-world conditions, cannot be estimated accurately using conventional tools, and are determined by the specifications of a protocol. This section examines the latency and throughput of communication channels in the proposed system, in addition to manual robot control.

\subsubsection{Communication Statistics}

\begin{table*}[th]
	\caption{Data transmission statistics for USB, Wi-Fi, and Bluetooth protocols.}
	\ra{1.3}
	\begin{center}
		\begin{tabular}{@{}llllll@{}}
			\hline
			Quantity & Buffer Size & USB Arduino  & USB V-USB & Wi-Fi & Bluetooth \\
					 & (Bytes)     & (@115200 bps) &           &       & \\
			\hline 
			Latency (ms) & 64 & 5.7 & 14.0 & 13.5 & 87.1 \\
			\hline
					    & 64 & 0.6 & 2.9 & 0.6 & 0.5 \\
			Throughput & 256 & 2.5 & 4.8 & 2.4 & 2.0 \\
			(KiBps)    & 512 & 5.0 & 5.4 & 4.8 & 3.9 \\
					  & 1024 & 9.9 & 5.8 & 9.5 & 6.7 \\
			\hline 
		\end{tabular}
		\label{tab:comm-stats}
	\end{center}
\end{table*}

Four communication types in the proposed system are the Arduino USB serial, software-based V-USB implementation for the custom AVR board, Wi-Fi, and Bluetooth. To test the latency of each network type, a packet containing the current timestamp identifier (a number) is sent from the Android device to the destination device (the microcontroller board or wireless server). This message is echoed back to the Android device to compare its arrival timestamp with the original message. This operation is repeated ten times for 100 timestamps (a total of 1000 messages), and the average timestamp difference is reported as the latency of the channel.

To measure throughput, 1000 packets of a predefined size are sent from the Android device to the target. Each packet contains a specific pattern of bytes. This packet is decoded in the target, and the number of successfully received bytes is sent to the Android device. The channel throughput is the number of successfully transmitted bytes divided by the duration of this test. The buffer size is increased for all modes of communication except the V-USB. For this configuration, the buffer size is kept at 64 bytes for all experiments because the current implementation of the proposed system cannot support buffers greater than or equal to 256 bytes. In this case, a train of $N_s$ consecutive 64-byte packets is sent to the target before a synchronous response for the number of received bytes is requested. Here, $N_s$ is the buffer size divided by 64, e.g., four in case of a buffer of 256 bytes.

For this experiment, an Android phone (Samsung Galaxy A20) establishes a Wi-Fi connection with a laptop controller server through an access point (Wi-Fi router) while it is directly paired with the desktop system for Bluetooth communication. Three types of Arduino devices considered here are Arduino Uno, Due, and Mega 2560. They are connected to the phone via a USB cable and operate at 115200 bits per second. The custom V-USB-powered AVR board is an Atmega32 with a clock frequency of 16 MHz.

Table \ref{tab:comm-stats} reports the results of this experiment. The Arduino board exhibits the lowest latency, while, as expected, Bluetooth communication demonstrates the highest value among these channels. The custom board almost has the same latency as the Wi-Fi connection and is in the mid-range. Although not explicitly shown in the table, the latency results for other buffer sizes are consistent. However, wireless communication shows a higher variance when changing the buffer size.

According to the table, throughput increases linearly with buffer size. The custom V-USB microcontroller has the highest throughput for small buffer sizes, while the Wi-Fi channel can provide high throughput for large buffers. The throughput of the Bluetooth connection is always lower than that of Wi-Fi, as expected. Also, the throughput of the V-USB board is consistently higher than that of the Arduino board for smaller buffers but changes slowly with an increase in buffer size. A future implementation of this board that increases the actual buffer size might improve the results. Based on these experiments, only the high-end and more expensive Arduino boards (such as Arduino Mega 2560) can provide enough memory for large buffer sizes.

In conclusion, the Arduino board is the preferred option for its low latency and high throughput using large buffer sizes in high-end boards, while the V-USB board is suitable for low-cost solutions. Moreover, when a Wi-Fi wireless connection is available, it is preferred over Bluetooth, as it offers high throughput and moderate latency.

\subsubsection{Manual Control of Robots}

\begin{figure}[ht]
	\centering
	\includegraphics[width=.40\textwidth]{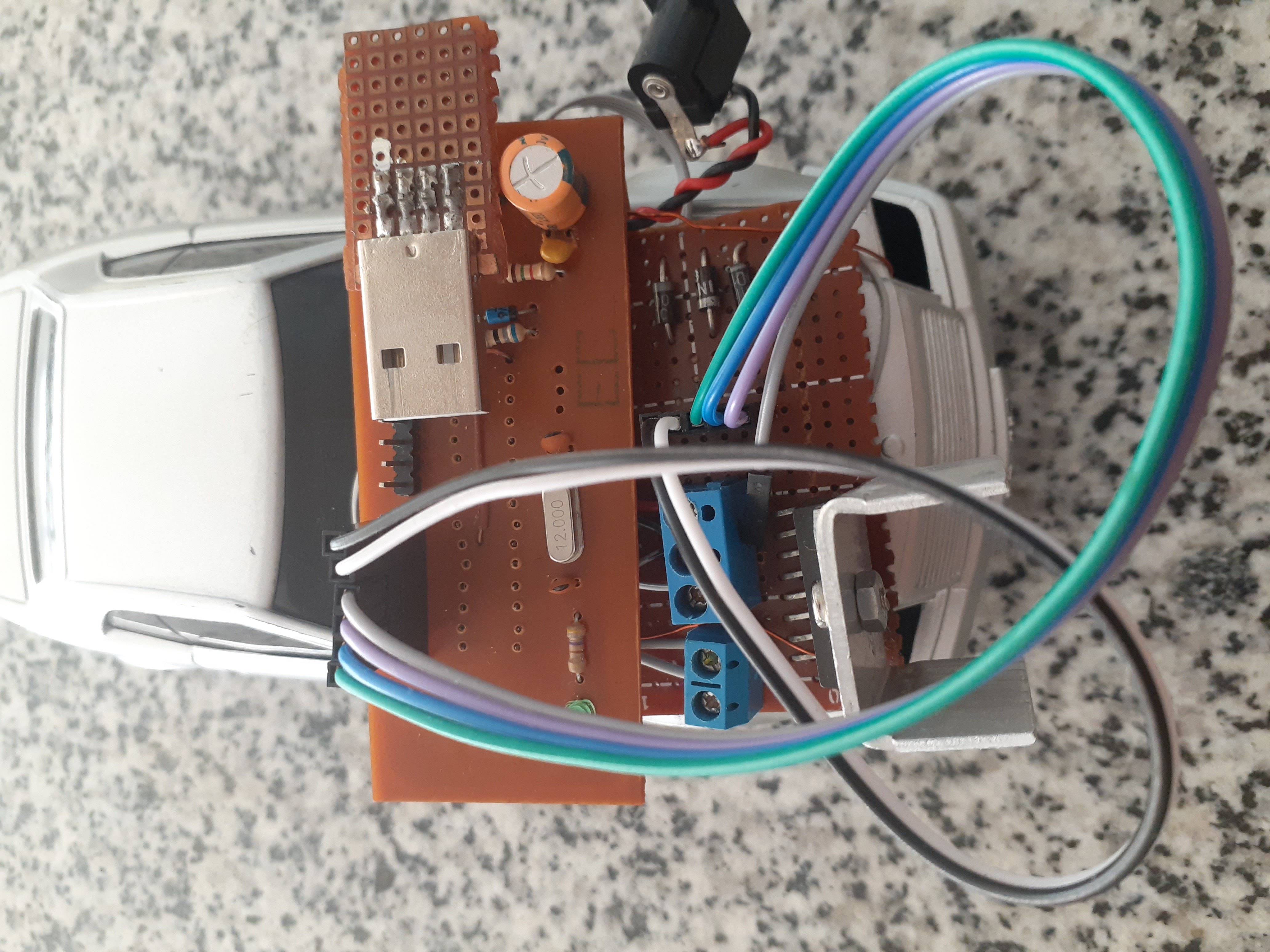}
	\caption{The toy car's setup for manual control using the proposed Robo-Platform system.}
	\label{fig:manual-toy-car-control}
\end{figure}

Finally, a ground vehicle and a quadcopter are manually controlled using the proposed system. In the first experiment, a Samsung Galaxy J5 device is attached to a V-USB-powered Atmega8 AVR board, which provides low-level control signals that are sent to a toy car's electrical DC motor through a driver (Fig. \ref{fig:manual-toy-car-control}). This car is controlled by two digital signals, one to enable/disable the device, and the other for forward/backward-rotation motion. These commands are issued wirelessly using the companion desktop application. The Android application processes the commands and sends the final driver signals to the microcontroller to be applied directly to the outputs%
\footnote{Video: \url{https://youtu.be/BTQ4yLB1bak}}.


Next, an Arduino board generates the PWM signals to control a quadcopter. The Android application receives the current state (the four PWM strength values) from the Arduino board, control commands from the wireless server, and IMU readings to calculate the device's orientation and update its state. These three inputs are combined to generate the desired PWM strength sent to the Arduino and directly applied to the device's PWM outputs. In both applications, the Android phone and microcontroller board are rigidly attached to the mobile robot and move with it. These experiments show that the proposed versatile platform can manually control various robotic systems.

\section{Conclusion} \label{sec:conclusion}

A robotic system for recording Android phones' sensors and controlling robots was proposed in this paper. In the data acquisition mode, the user can select a combination of available internal sensors and ADC measurements from an external USB-powered microcontroller board. After providing the necessary permissions, the user can record sensors in a raw format suitable for 3D reconstruction and pose estimation applications. The experiments show that the recorded dataset can be fed into modern SLAM systems, though more robust algorithms are required to handle the noisy and sporadic sensor data from an Android device.

In the robot control mode, the Android component of the proposed system can receive control commands from a remote desktop or Android server via Bluetooth or Wi-Fi, process these commands, and send the resulting control signals to either an Arduino or custom V-USB microcontroller board via USB. Since the reliability of communication links is crucial in this application, the latency and throughput statistics have been provided for various types of connectivity. Although Wi-Fi supports high throughput, the Arduino board also offers low latency. Moreover, the proposed system successfully applied to manual control of a miniature toy car and a quadcopter.

The future versions of the proposed platform can acquire data from various external USB or wireless sensors. Additionally, the Android application can incorporate improved real-time data visualization. In the robot control mode, provisions for sending sensor data, including images from the Android device, to the remote server for surveillance applications can be considered. Furthermore, it is possible to integrate SLAM/AR algorithms with the proposed system to offer online state estimation for entertainment or self-navigating agents.

%
%

\bibliographystyle{ieeetr}
\bibliography{References}

\end{document}